\documentclass{article}     
\usepackage{times}
\usepackage{soul}
\usepackage{url}
\usepackage[utf8]{inputenc}
\usepackage[small]{caption}
\usepackage{amsmath}
\usepackage{amsthm}
\usepackage{booktabs}
\usepackage{algorithm}
\usepackage{algorithmic}
\usepackage[algo2e]{algorithm2e}
\usepackage{verbatim}
\usepackage{graphicx}
\usepackage{subfigure}
\usepackage{amsmath}
\usepackage{bm}
\usepackage{booktabs}
\usepackage{color}
\usepackage{amsmath,epsfig,bm, amssymb,subfigure,graphicx,algorithm,algorithmic,color}
\newtheorem{defi}{Definition}

\newtheorem{lemm}[defi]{Lemma}
\newtheorem{theo}[defi]{Theorem}

\newcommand{\theoproof}[1]{\noindent{\bf (Proof of Theorem #1) }}
\newcommand{\lemmproof}[1]{\noindent{\bf (Proof of Lemma #1) }\\}
\newcommand{\proofend}{\hfill$\Box$\vspace{2mm}}

\newcommand{\argmin}{\mathop{\mathrm{argmin\,}}}

\newcommand{\inner}[2]{\langle #1,#2\rangle}

\newcommand{\mathbbR}{\mathbb{R}}

\newcommand{\boldone}{{\boldsymbol{1}}}

\newcommand{\boldA}{{\boldsymbol{A}}}

\newcommand{\boldC}{{\boldsymbol{C}}}

\newcommand{\boldG}{{\boldsymbol{G}}}
\newcommand{\boldH}{{\boldsymbol{H}}}
\newcommand{\boldI}{{\boldsymbol{I}}}

\newcommand{\boldP}{{\boldsymbol{P}}}

\newcommand{\boldR}{{\boldsymbol{R}}}

\newcommand{\boldW}{{\boldsymbol{W}}}

\newcommand{\boldb}{{\boldsymbol{b}}}

\newcommand{\bolde}{{\boldsymbol{e}}}

\newcommand{\boldk}{{\boldsymbol{k}}}

\newcommand{\boldp}{{\boldsymbol{p}}}

\newcommand{\boldx}{{\boldsymbol{x}}}
\newcommand{\boldy}{{\boldsymbol{y}}}
\newcommand{\boldz}{{\boldsymbol{z}}}

\newcommand{\bolddelta}{{\boldsymbol{\delta}}}

\newcommand{\boldphi}{{\boldsymbol{\phi}}}

\newcommand{\ntr}{n}

\newcommand{\norm}[1]{\left\lVert#1\right\rVert}

\DeclareMathOperator*{\loss}{loss}
\DeclareMathOperator*{\optloss}{reg}

\DeclareMathOperator*{\tra}{Tr}

\usepackage{color}

\advance\oddsidemargin-1in
\usepackage{graphicx}
\usepackage{subfigure}
\usepackage{algorithm}
\usepackage{algorithmic}
\usepackage{subfigure}
\usepackage{multirow}

\usepackage{enumitem}
\setlist{nolistsep}

\usepackage{array}
\usepackage{amsmath}
\usepackage{bm,amssymb,color,amsthm}
\usepackage{tabularx}
\usepackage{booktabs}

\title{Fast local linear regression with anchor regularization}

\author{
Mathis Petrovich$^{1,2}$\footnote{Work done at RIKEN AIP},
Makoto~Yamada$^{2,3}$\\
$^1$ENS Paris-Saclay, $^2$RIKEN AIP, $^3$Kyoto University\\
 mathis.petrovich@ens-paris-saclay.fr, makoto.yamada@riken.jp}

\begin{document}

\maketitle

\begin{abstract}
    Regression is an important task in machine learning and data mining. It has several applications in various domains, including finance, biomedical, and computer vision. Recently, network Lasso, which estimates local models by making clusters using the network information, was proposed and its superior performance was demonstrated.  
    In this study, we propose a simple yet effective local model training algorithm called the fast anchor regularized local linear method (FALL). More specifically, we train a local model for each sample by regularizing it with precomputed anchor models. The key advantage of the proposed algorithm is that we can obtain a closed-form solution with only matrix multiplication; additionally, the proposed algorithm is easily interpretable, fast to compute and parallelizable. Through experiments on synthetic and real-world datasets, we demonstrate that FALL compares favorably in terms of accuracy with the state-of-the-art network Lasso algorithm with significantly smaller training time (two orders of magnitude).
\end{abstract}

\section{Introduction}
The regression problem is an important problem in machine learning, data mining, and statistics, and several research works have investigated it in the past decades. Examples include stock price prediction \cite{finance2,finance1}, age prediction from RNA-seq \cite{agepred} or images \cite{agepredimg}, sentimental analysis \cite{sentiment2,sentiment1}, or house prediction \cite{housepred} to name a few.

The most widely used regression approach is based on a linear model including the ordinary least squares (OLS), Ridge regression, least absolute shrinkage and selection operator (Lasso) \cite{lasso}, and elastic net \cite{zou2005regularization}. Because these linear models are extremely simple and can be interpreted by simply checking the linear coefficients of the variables; these approaches are in particular used in practice. However, one of the limitations of linear models is that they cannot handle complex nonlinear data; the performance can be significantly degraded if we apply these linear methods to process complex data such as the gene expression data used heavily in biology and healthcare.

To handle complex data, researchers tend to use kernel methods such as kernel ridge regression (KRR) and support vector regression (SVR) \cite{book:Schoelkopf+Smola:2002}. These approaches are nonlinear models and can handle more complex relationships between inputs and outputs.  However, because they are global methods, the output is somewhat smooth, which can be prejudicial for highly complex data with some hard jumps (see Figure \ref{fig:simple_a}). Moreover, in general, it is difficult to interpret the features in kernel methods.

Instance-based learning such as the k-nearest neighbor (KNN) algorithm \cite{knn} is also a type of widely used nonlinear model. KNN can be considered a local method, which uses the distance or similarity function to predict the output. Despite their good prediction ability, KNN methods can undergo extrapolation when a new input is significantly distant from the training data and can also overfit to much on the data. 
 Moreover, because KNN is a non-parametric method, it is generally difficult to interpret the model.

Recently, the network Lasso, which is a novel type of local method was proposed \cite{hallac2015network,hallac2015snapvx}. Specifically, it estimates the local models by simultaneously clustering the samples and estimating the linear model from them. Accordingly, it combines local regression with network-based sparse regularization (a.k.a., generalized fused regularizer). Because the network Lasso is a locally linear model, it can be easily interpreted to be similar to linear models. More recently, its extension to high-dimensional data with exclusive regularizer has also been proposed \cite{yamada2016localized}. The network Lasso outperforms existing nonlinear models in prediction accuracy, and it was demonstrated that it can automatically cluster models. However, the nearest neighbor graphs should be pre-computed before training the network Lasso. Moreover, the computation of network regularization is expensive and it does not scale to million data points.

In this study, we propose a simple yet efficient local regression model called the fast anchor regularized local linear (FALL) model. Specifically, we estimate a model for each sample, where the model parameter is regularized to be similar to the pre-computed models (called anchor models). The key advantage of the proposed FALL model is that it has a closed-form solution and it is extremely faster than network Lasso. Moreover, the FALL model can cluster samples by choosing the anchor models.
 Finally, it can extrapolate even when the data is far from the input data owing to the locally linear model. Through experiments, it was found that the proposed algorithm compares favorably with existing state-of-the-art algorithms in accuracy, and it can be 500 times faster for training than network Lasso.
 
 \vspace{.05in}
\noindent {\bf Contribution:} The contributions of this study are summarized below.
\begin{itemize}
    \item We propose a new locally linear regression problem with anchor regularization.
    \item We derive the closed-form solution of the anchor regularization problem.
    \item We extensively investigate the proposed method to determine when it worked. 
\end{itemize}

\section{Related Work}
In this section, we review the linear and nonlinear regression methods.

\vspace{.05in}
\noindent {\bf Linear methods:} Linear models are the most widely used regression model. Ridge regression, the least absolute shrinkage and selection operator (Lasso) \cite{JRSSB:Tibshirani:1996}, and the elastic-net \cite{zou2005regularization} rely on the following modeling:
\[
\boldy_i =
\boldW^\top \boldx_i + \bolde_i
\]
where $\boldW \in \mathbbR^{d \times m}$ is the model parameter and $\bolde_i$ is the noise vector of the $i$-th sample. Because the model only depends on $\boldW$ for all samples, it is called a \emph{global} model. 
These methods differ on the basis of regularizing the weight parameter $\boldW$. The linear model is the first choice for data analysis and it can be scaled efficiently for large dataset. However, it tends to demonstrate poor performance if the data are complex (i.e., expression data in biology).

\vspace{.05in}
\noindent {\bf Nonlinear methods:} If the data are complex, nonlinear models can be used. Kernel ridge regression (KRR) can solve the same linear problem as ridge regression, but it maps the original vector $\boldx_i$ into a reproducing kernel Hilbert space (RKHS) with a mapping $\boldphi$ \cite{book:Schoelkopf+Smola:2002}. Specifically, KRR is modeled as follows. 
\[\boldy_i = \boldW^\top \boldphi(\boldx_i) + \bolde_i = \widetilde{\boldW}\boldk(\boldx_i)+ \bolde_i,
\]
where $\widetilde{\boldW} \in \mathbbR^{n \times n}$ is the model parameter, $\boldk(\boldx) = (K(\boldx_1, \boldx), K(\boldx_2, \boldx), \ldots, K(\boldx_n, \boldx))^\top \in \mathbbR^{n}$, and $K(\boldx, \boldx') = \boldphi(\boldx)^\top \boldphi(\boldx')$ is a kernel function such as polynomial or RBF kernel. The kernel methods rely on the choice of kernel function.  KRR is a simple yet effective nonlinear method. However, owing to the nonlinear transform, it is difficult to interpret KRR. Moreover, the KRR output tends to be over smoothed due to the kernel function.

KNN is also a widely used nonlinear regression model that predicts the average of $\boldy_i$ of the nearest neighbors for the input $\boldx$. KNN tends to work for interpolation problems, while it is not suited for extrapolation. Moreover, for KNN, the output performance tends to be affected by the number of nearest neighbors. 

Recently, the network Lasso was proposed \cite{hallac2015network}, as another local linear model. The optimization problem of network Lasso is given as
\[
\min_{\{\boldW\}_{i = 1}^n}~\sum_{i = 1}^n \|\boldy_i - \boldW_i^\top \boldx_i\|_2^2 + \lambda \!\!\!\sum_{(j, k) \in E}\!\! r_{j,k} \norm{\boldW_j - \boldW_k}_F,
\]
where $r_{j,k} \geq 0$ is the pre-defined graph information, $\|\cdot\|_2$ is the $\ell_2$ norm, $\|\cdot\|_F$ is the Frobenius norm, and $\lambda \geq 0$ is the regularization parameter. 
In network Lasso, owing to the fused Lasso type regularizer, as $\lambda$ increases, the models tend to form a consensus. Thus, we can automatically cluster the model parameters. However, the optimization of network Lasso is expensive and it does not scale well to large datasets. 
 
\section{Problem Formulation}
\label{sec:prob}
In this section, we formulate the locally linear regression problem. 

Let us denote an input vector by $\boldx = (x^{(1)}, \dots, x^{(d)})^\top \in \mathbbR^d$ and the corresponding output value by $\boldy \in \mathbbR^m$. The set of training samples $\{(\boldx_i, \boldy_i)\}_{i = 1}^{\ntr}$ has been drawn i.i.d. from a joint probability density $p(\boldx, \boldy)$. In this study, we consider a local linear model
\[
\boldy_i = \boldW_i^\top \boldx_i + \bolde_i,
\]
where $\boldW_i \in \mathbbR^{d \times
  m}$ is the $i$-th sample regression parameter and $\bolde_i \in \mathbbR^{m}$ is the noise vector of the $i$-th sample. Note that each $\boldy_i$ is
modeled by its model parameter $\boldW_i$, while a \emph{global} model employs $\boldW$ to model all $\boldy_i$.


The study aims to estimate $\boldW_i$ from the training input-output samples efficiently. This is intrinsically a very difficult problem because we need to estimate $\boldW_i$ from one sample. That is, it can easily be overfitted to the training sample. Thus, a regularizer that can efficiently generalize to an unknown sample is the key to this problem.

\section{Proposed Method}
In this section, we propose the FALL method.

\subsection{Fast anchor regularized local linear  method}
For a training sample pair $(\boldx_i, \boldy_i)$, we consider the following  regression problem:
\begin{align}
\label{eq:main_obj}
    \min_{\boldW_i, \boldp_i} &~~ J_i(\boldW_i, \boldp_i),~~\text{s.t.}~~ \boldp_i \in S_k,
\end{align}
where 
\begin{align*}
    J_i(\boldW_i, \boldp_i) &= \norm{\boldy_i - \boldW_i ^\top \boldx_i}_2^2 + \lambda\sum_{l = 1}^{k} p_{il}\|\boldW_i - \boldA_l\|^2_F
\end{align*}
and
\begin{displaymath}
    S_k = \{ \boldp \in \mathbbR_+^k \text{ s.t. }
    \boldp^\top \boldone_k = 1 \} 
\end{displaymath}
is the simplex set, $\boldone_k \in \mathbbR^{k}$ is the vector whose elements are all one, $\boldA_l \in \mathbbR^{d \times m}$ is the pre-defined model parameter (we call this anchor model), $k \ll n$ denotes the number of anchors, and $\lambda > 0$ is the regularization parameter. $\boldp_i$ is the parameter that represents the contribution of each anchor model in the regularization.

Note that a bias term can be used for every model $\boldW_i$. To accomplish this, the bias should be used for computing the anchor models $\boldA_1, \ldots, \boldA_k$. Then, each data point $\boldx_i \in \mathbbR^d$ must be replaced by $(\boldx_i^\top, 1)^\top \in \mathbbR^{d+1}$.



\begin{figure}[t]
\begin{center}
   \includegraphics[scale=0.8]{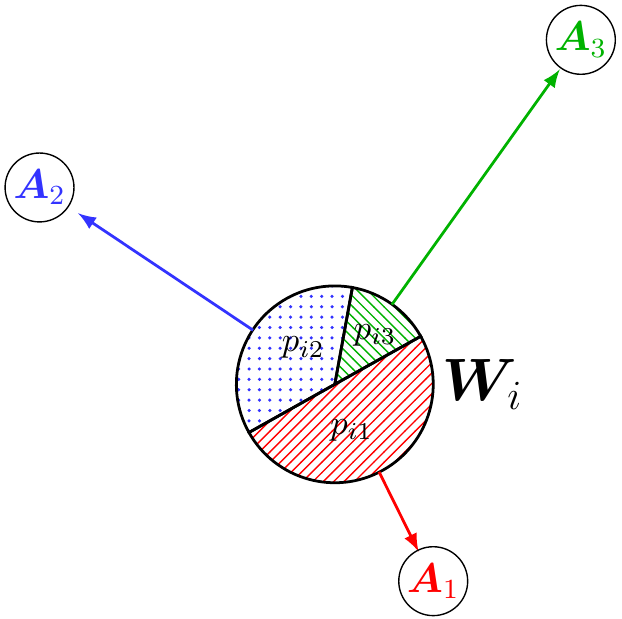}
\end{center}
    \caption{Representation of models in the parameter space. The area of the sectors is represented by the vector $\boldp_i$, which represents the contribution of each $\boldA_l$ in the regularization. $\boldp_{i}$ represents how $\boldW_i$ is shared among all anchor models. The optimization of our model attempts to provide $\boldW_i$ a good power of prediction (to get $\boldy_i$) but $\boldW_i$ should not be too far from the anchors because the regularization penalizes the distance of $\boldW_i$ to the $\boldA_l$. Simultaneously, the optimization must choose $\boldp_i$, which determines the importance of $\boldW_i$ being close to each $\boldA_l$ because when $p_{il}$ is large, the cost of being further from $\boldA_l$ is large. The arrows represent the sharing between the model and the anchors.}
    \label{fig:pic1}
\end{figure}

The anchor parameters play the role of "reference" models. This can be a priori models or can be constructed with local models (see section \ref{sec:anchors}).
They are not optimized in this regularizer; they are known beforehand.
The idea of this regression problem is to jointly optimize
\begin{itemize}
\item The data attachment term (the square error term) that attempts to efficiently model $\boldy_i$ from $\boldx_i$.
\item The anchor attachment term (the convex sum of squared differences), which can be interpreted as "sharing" (see Figure \ref{fig:pic1}), that attempts to force the model $\boldW_i$ to be as similar to the anchor model $\boldA_l$.
\end{itemize}
As the first term does not depend on $\boldp_i$, its minimization can be put in the second term; thus, the whole problem can be observed as the one that finds the best model $\boldW_i$, which minimizes the data term and the \emph{best} convex sum on the right. In this sense, the second term with the minimization over $\boldp_i$ is the regularizer of our problem. The best $\boldp_i$ can be interpreted as weights, which force the model $\boldW_i$ to be close to the "good" anchor models, where their predictions, given $\boldx_i$ are close to $\boldy_i$.

This regularizer is a relaxation of the optimal transport (OT) \cite{peyre2019computational} setting because the sum of all values of $i$ yields the same formula. The main difference is that we decided to not constrain each anchor model to share its information with the models by a fixed amount (as the second constraint in the OT). Indeed, in some data structures, two anchor models can have different impacts, and one can have a larger influence than the other (see Figure \ref{fig:simple_c} where more points are better represented by the blue model) and thus, it is not profitable to restrict an anchor model to share its power over all models $\boldW_i$. That is, $\boldp_i$ can be used to softly cluster $\boldx_i$ with respect to the anchor models.

A standard approach to solve the optimization problem Eq.~\eqref{eq:main_obj} is to alternatingly minimize $\boldW_i$ and $\boldp_i$. However, it can be trapped by a poor local optimal solution. In this paper, we derive the analytical solution of the optimization problem Eq.~\eqref{eq:main_obj}.

\subsection{Minimization of weights given the contributions of anchor models}
We first derive the optimal solution of $\boldW_i$ by fixing $\boldp_i$. Because the function to optimize $\boldW_i \rightarrow J_i(\boldW_i, \boldp_i)$ is convex, we can cancel the derivative with respect to $\boldW_i$ to determine the optimal $\boldW_i$:
\begin{align*}
  \frac{\partial J}{\partial \boldW_i} (\boldW_i, \boldp_i) &= 2 \boldx_i (\boldx_i ^ \top \boldW_i  - \boldy_i ^ \top) + 2 \lambda \sum_{l=1}^k p_{il} (\boldW_i - \boldA_l) = 0.
\end{align*}
Thus, the optimal solution is given by
\begin{displaymath}
   \widehat{\boldW}_i = \left( \boldx_i \boldx_i ^ \top + \lambda \boldI_d \right)^{-1}(\boldx_i \boldy_i^\top + \lambda \boldG_i),
\end{displaymath}
where $\boldG_i = \sum_{l=1}^k p_{il} \boldA_l \in \mathbbR^{n \times m}$ is the mixture of anchor models.
We can obtain the optimal solution of $\boldW_i$ analytically. However, because we need to compute it for all samples, it is computationally expensive if the number of features is extremely large (e.g., $d = 10^5$).

Owing to the effective structure of the inverse matrix (i.e., rank one matrix plus an identity matrix), we can use the Sherman-Morrison formula \cite{sherman1950,bartlett1951} (or simple algebra) to obtain the analytical solution without inverting the matrix.
\begin{displaymath}
  (\boldx_i \boldx_i ^ \top + \lambda \boldI_d)^{-1} = \frac{1}{\lambda} \boldI - \frac{1}{\lambda (\lambda + \boldx_i ^ \top \boldx_i)} \boldx_i \boldx_i ^ \top.
\end{displaymath}
Finally, we have
\begin{displaymath}
    {\widehat{\boldW}_i = \frac{1}{\lambda + \boldx_i ^\top \boldx_i} \boldx_i \left[\boldy_i  - \boldG_i ^\top \boldx_i \right] ^\top + \boldG_i }.
\end{displaymath}

\noindent Because the final solution does not include the inverse matrix, it can be estimated easily and quickly. 

\subsection{Minimization of the global loss}

In this section, we minimize $\boldp_i \in S_k \rightarrow J_i(\widehat{\boldW}_i, \boldp_i)$ by taking the best $\widehat{\boldW}_i$ computed in the last section. More specifically, we derive the global optimal solution of $\boldp_i$.

\noindent Thus, we plug the optimal $\widehat{\boldW}_i$ into Eq.~\eqref{eq:main_obj}, and consider the following optimization problem:
\begin{align}
\label{eq:eq2}
    \min_{\boldp_i} &~~ J_i(\widehat{\boldW}_i, \boldp_i),~~\text{s.t.}~~ \boldp_i \in S_k.
\end{align}

\begin{lemm}\label{lemm:quad}
Eq.~\eqref{eq:eq2} is equivalent to the following quadratic problem:
\begin{align*}
    \min_{\boldp_i} &~~ \boldp_i ^ \top \boldH_i \boldp_i + \boldb_i ^\top \boldp_i + \beta_i \boldy_i ^\top \boldy_i,~~\text{s.t.}~~ \boldp_i \in S_k,
\end{align*}
where
\begin{itemize}
\item $\beta_i = \lambda/(\lambda + \norm{\boldx_i}_2^2)$.
\item $\boldH_i \in \mathbbR^{k \times k}$ such that
  $(\boldH_i)_{ll^{\prime}} = \beta_i \boldx_i ^\top \boldA_l \boldA_{l^{\prime}}
  ^\top \boldx_i - \lambda \tra(\boldA_l ^\top
  \boldA_{l^{\prime}})$. $\tra(\cdot)$ is the trace operator.
\item $\boldb_i \in \mathbbR^{k}$ such that $(\boldb_i)_l = - 2 \beta_i \boldy_i ^ \top \boldA_l ^\top \boldx_i + \lambda \norm{\boldA_l}^2_F$.
\end{itemize}
\end{lemm}

\noindent The proof can be found in the appendix. It is based on the rewriting of $J_i(\widehat{\boldW}_i, \boldp_i)$ with some factorization.

\begin{lemm}\label{lemm:concave}
 $\boldH_i$ is symmetric negative definite, and Eq.~\eqref{eq:eq2} is a concave QP under linear constraints.
\end{lemm}

\lemmproof{\ref{lemm:concave}} 
First, by definition, $\boldH_i$ is symmetric.
Let $\boldz \in \mathbbR^k$, and let us show that $\boldz ^\top
\boldH_i \boldz < 0$.
\begin{align*}
  \boldz ^\top \boldH_i \boldz &= \beta_i \sum_{l=1}^{k} \sum_{l^{\prime}=1}^{k} z_{il} z_{il^{\prime}} (\boldx_i ^\top \boldA_l \boldA_{l^{\prime}} ^\top \boldx_i)  -\lambda \sum_{l=1}^{k} \sum_{l^{\prime}=1}^{k} z_{il} z_{il^{\prime}} \tra(\boldA_l^\top \boldA_{l^{\prime}}) \\
                               &= \frac{\lambda}{(\lambda + \boldx_i ^\top \boldx_i)} \norm{\sum_{l=1}^{k}\! z_{il} \boldA_l^\top \boldx_i}_2^2 \!-\! \lambda \norm{\sum_{l=1}^{k}\! z_{il} \boldA_l}^2_F \\
                               &\leq \lambda \norm{\sum_{l=1}^{k} z_{il} \boldA_l}^2_F \left( \frac{\boldx_i ^\top \boldx_i}{\lambda + \boldx_i ^\top \boldx_i} - 1 \right) \\ 
                               & < 0
\end{align*}
The first inequality is valid owing to the submultiplicativity of the Frobenius norm. 
 Thus, our problem is a strictly concave problem. \proofend

\begin{theo}
\label{th:final}
    A minimizer of our regression problem Eq.~\eqref{eq:main_obj} can be expressed as
    \begin{equation}
    \label{eq:finalsol}
        \widehat{\boldW}_i = \frac{1}{\lambda + \boldx_i ^\top \boldx_i} \boldx_i \left[\boldy_i  - \boldA_{l_i} ^\top \boldx_i \right] ^\top + \boldA_{l_i}
    \end{equation}
    where $l_i = \argmin_{l} \norm{\boldy_i - \boldA_l ^\top \boldx_i}_2^2$
\end{theo}

\theoproof{\ref{th:final}} Owing to the two lemmas (\ref{lemm:quad} and \ref{lemm:concave}), we know that the function $\boldp_i \in S_k \rightarrow J(\widehat{\boldW_i}, \boldp_i)$ is concave. This function is also a continuous function defined on the set $S_k$, which is convex and compact. By the Bauer minimum principle, this function attains its maximum at some extreme point of that set.

The boundary of the simplex $S_k$ is the set of points $\boldp_i$ such that there is one coordinate with a $1$ and all others are set to $0$. This implies a hot vector that indicates an anchor model.

\begin{figure}[t]
\begin{center}
 \includegraphics[scale=0.8]{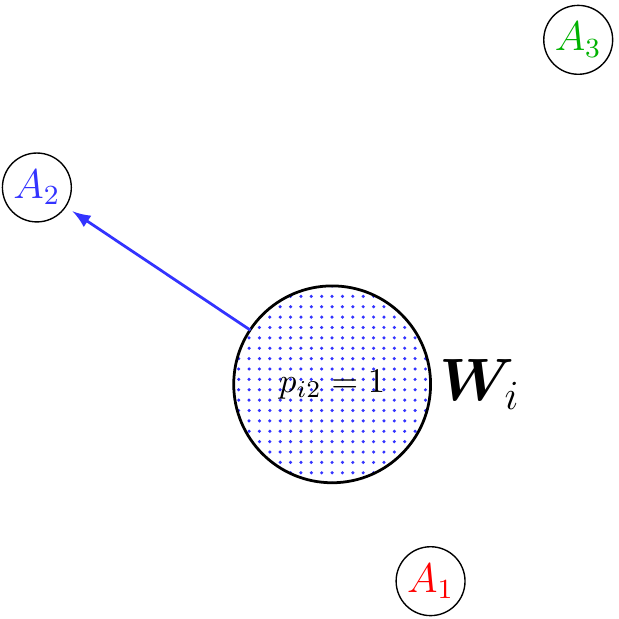}
\end{center}
    \caption{In the optimal solution, $\boldW_i$ is not shared but it chooses a unique anchor model $\boldA_l$, which effectively fits $\boldx_i$ to $\boldy_i$ (i.e., minimizes $|| \boldy_i - \boldA_l ^\top \boldx_i||^2_2$). The best anchor model could be $\boldA_2$ in blue so that the optimal $\boldp_i$ is the hot vector with $1$ in position $l=2$.} 
    \label{fig:pic2}
\end{figure}

\noindent Plugging the solution in $J$ and simplifying it, we obtain
\begin{align*}
J(\bolddelta_l, \widehat{\boldW}_i) &= \beta_i \left( \norm{\boldA_l ^\top \boldx_i}_2^2 - 2 \inner{\boldy_i}{\boldA_l ^\top \boldx_i}_2 + \boldy_i ^\top \boldy_i \right)\\
&= \beta_i \norm{\boldy_i - \boldA_l ^\top \boldx_i}_2^2,
\end{align*}
which is the problem of finding the best anchor model for the data point $\boldx_i$ (see Figure \ref{fig:pic2}). This can be found by testing all possibilities, i.e.,
\begin{displaymath}
    p_{il}^* = \begin{cases}
            1 \text{ if } l = \argmin_{l} \norm{\boldy_i - \boldA_l ^\top \boldx_i}_2^2 \\
            0 \text{ otherwise.}
            \end{cases}.
\end{displaymath}
By substituing this into $\widehat{\boldW}_i$, $\boldG_i^*$ becomes $\boldA_{l_i}$ which shows that it is not a mixture but an anchor model (the one that predicts the closest to $\boldy_i$ given $\boldx_i$). \proofend

Thanks to Theorem \ref{th:final}, the solution can be computed easily and quickly. The optimal solution might not be unique if there exists $\boldA_{l_1}$ and $\boldA_{l_2}$ such that $\boldA_{l_1} ^\top \boldx_i = \boldA_{l_2} ^\top \boldx_i$. In this situation, either can be chosen.

It is interesting that the global minimizer for the assignment $\boldp_i$ does not depend on $\lambda$ but only depends on the power of prediction of each anchor model to predict something close to $\boldy_i$ given the data point $\boldx_i$.

\begin{algorithm}[t]
  \caption{Local linear regression with anchor regularization.\label{alg:all}}
  \begin{algorithmic}
    \STATE {\bfseries Input:} $(\boldx_1, \ldots,
  \boldx_n)$, $(\boldy_1, \ldots, \boldy_n)$
    \STATE $\boldA_1, \ldots, \boldA_k \gets$ Finding anchors by Algorithm \ref{alg:anc}
    \FOR{$i \gets 1,\, n$}
    \STATE $l_i \gets \argmin_l \norm{\boldy_i - \boldA_l^\top \boldx_i}$
    \STATE $\boldp_i \gets \bolde_{l_i}$
    \STATE $\boldW_i \gets \frac{1}{\lambda + \boldx_i ^\top \boldx_i} \boldx_i \left[\boldy_i  - \boldA_{l_i} ^\top \boldx_i \right] ^\top + \boldA_{l_i}$
    \ENDFOR
    \STATE {\bfseries Output:} $(\boldW_1, \ldots, \boldW_n), (\boldp_1 , \ldots, \boldp_k)$
  \end{algorithmic}
\end{algorithm}

Given the anchor models, Algorithm \ref{alg:all} can be efficiently computed even if the dimension of $d$ is large. 
Note that the entire Algorithm \ref{alg:all} is embarrassingly parallel and therefore can be vectorized to speed up the process. Moreover, given the simple matrix form, the proposed method can be implemented on a GPU.


\subsection{Prediction}
In this section, we will focus on the prediction power of each model $\widehat{\boldW}_i$ of our system. We will use the training data at first, then with new samples and finally, we will see how to combine all the models and how to interpolate between them.

\vspace{.05in}
\noindent {\bf Prediction with the training data:} 
The prediction $\widehat{\boldy}_i = \widehat{\boldW}_i^\top \boldx_i$ can be decomposed as
\begin{equation}
\label{eq:opt_Wi}
  \widehat{\boldW}_i^\top \boldx_i = \underbrace{ \frac{\boldx_i^\top \boldx_i}{\lambda + \boldx_i ^\top \boldx_i}}_{1 - \beta_i} \boldy_i + \underbrace{\frac{\lambda}{\lambda + \boldx_i ^\top \boldx_i}}_{\beta_i}  \boldA_{l_i}^\top \boldx_i,
\end{equation}
 where $0<\beta_i <1$ controls the contribution of the anchors to the prediction.

\noindent That is, the prediction $\widehat{\boldW}_i^\top \boldx_i$ is a convex
sum of the ground truth $\boldy_i$ and the prediction of the best anchor model $\boldA_{l_i}^\top \boldx_i$. 

\noindent The $\lambda$ parameter controls the regularization as follows.
\begin{itemize}
\item When $\lambda$ is small ($\lambda \ll \boldx_i^\top
  \boldx_i$), the coefficient $\beta_i \simeq 0$, which yields a prediction close to the ground truth. This corresponds to an unregularized local linear model.
\item When $\lambda$ is large ($\lambda \gg \boldx_i^\top \boldx_i$), the coefficient $\beta_i \simeq 1$, which implies that the prediction is close to the prediction of the best anchor model and not much emphasis is given on its value being close to $\boldy_i$. 
\end{itemize}
This provides a way of interpreting the role of $\lambda$.
The training error made by our model always satisfies
\begin{displaymath}
  \epsilon_i = \norm{\boldy_i - \widehat{\boldy}_i}_2 = \beta_i \norm{\boldy_i - \boldA_{l_i}^\top \boldx_i}_2,
\end{displaymath}
which is bounded by the minimal error of the anchor models.
The optimal model parameter can be decomposed as
\begin{displaymath}
\widehat{\boldW}_i = \boldC_i + \boldA_{l_i}
\end{displaymath} 
where
\begin{displaymath}
\boldC_i = \frac{1}{\lambda + \boldx_i ^\top \boldx_i} \boldx_i \left[\boldy_i  - \boldA_{l_i} ^\top \boldx_i \right] ^\top
\end{displaymath}
can be interpreted as a "correction" model because $\boldC_i^\top \boldx_i = (1 - \beta_i) (\boldy_i - \boldA_{l_i}^\top \boldx_i)$. This correction term is added to the prediction of the anchor model to traverse from the full regularized prediction $\boldA_{l_i} ^\top \boldx_i$ (when $1 - \beta_i = 0$ where $\lambda \rightarrow \infty$) to the non regularized prediction $\boldy_i$ (when $\beta_i = 0$ where $\lambda \rightarrow 0$). The correction model is the model that adds to the anchor model to adjust the prediction to get closer to the ground truth.

\vspace{.05in}
\noindent {\bf Prediction with new samples:}
Let us compute the prediction of model $\widehat{\boldW}_i$ with a new input $\boldx$:
\begin{displaymath}
    \widehat{\boldW}_i  ^\top \boldx = \underbrace{\frac{\boldx_i ^\top \boldx}{\lambda + \boldx_i ^\top \boldx_i}
    \left[\boldy_i  - \boldA_{l_i} ^\top \boldx_i \right]}_{\boldC_i ^\top \boldx} + \boldA_{l_i}^\top \boldx.
\end{displaymath}
We can notice interesting properties in this prediction.
\begin{itemize}
    \item When $\boldx$ is very different from $\boldx_i$, i.e., when $|\boldx_i ^\top \boldx|$ is very small, the correction term $\boldC_i ^\top \boldx$ is very small; thus, this model does not take any risks and behaves like the anchor model.
    \item When $\boldx$ and $\boldx_i$ are highly correlated (i.e., $|\boldx_i ^\top \boldx|$ is large), the correction term has more importance. The term is proportional to $\boldx_i ^\top \boldx$ to adjust the correction in the good direction (as it can be negative).
    \item $\lambda$ is used to adjust the regularization: when $\lambda$ is large, the prediction of this model will always be close to the prediction of the best anchor model.
    \item If $\boldx_i ^\top \boldx_i = 0$, i.e., $\boldx_i = 0$, it is not possible to obtain a $\boldy_i \neq 0$ with a linear model. In this case, the prediction will be based on the anchor model.
\end{itemize}

\vspace{.05in}
\noindent {\bf Combining the models:}
After fitting all the models $\boldW_i$, we determine how to use them to predict the final prediction $\widehat{\boldy}_i$ with a new vector $\boldx$. More specifically, we do not know the corresponding parameter of the new input $\boldx$. Thus, we propose a simple yet effective prediction approach.

Our assumption is that locally, the data behave like linear models. Accordingly, let us compute the $K_p \geq 1$ nearest neighbors $\boldx_{h_1}, .., \boldx_{h_{K_p}}$ for $\boldx$ and determine the corresponding models $\boldW_{h_1}, .., \boldW_{h_{K_p}}$.

Each model $\boldW_{h_j}$ is locally linear and can be used to predict $\boldW_{h_j}^\top \boldx$. 
The models created with a vector close to $\boldx$ should yield better results. Considering this, a weighted average of the local predictions can be predicted by
\begin{align}
\label{eq:eq3}
    \boldy_{\text{pred}} &= \sum_{j=1}^{K_p} \alpha_{h_j} \boldW_{h_j}^T \boldx = \underbrace{\left[\sum_{j=1}^{K_p} \alpha_{h_j} \boldW_{h_j}\right]^\top}_{\boldW_{\text{pred}}^\top} \boldx,
\end{align}
where the weights $\alpha_{h_1}, .., \alpha_{h_{K_p}}$ are inversely proportional to the distance from $\boldx$ to $\boldx_{h_j}$:
\begin{align*}
    \alpha_{h_j} = \frac{ \frac{1}{d(\boldx, \boldx_{h_j})} }{ \sum_{j=1}^{K_p} \frac{1}{d(\boldx, \boldx_{h_j})} },
\end{align*} 
with the distance $d(\boldx, \boldy) = \norm{\boldy - \boldx}_2$. \\

For each $\boldx$, our algorithm (in Algorithm \ref{alg:pred}) predicts the local average model $\boldW_{\text{pred}}$. Specifically, Algorithm \ref{alg:pred} first computes the $K$ nearest neighbors of $\boldx$ in the training set. Then, it computes $\boldW_{\text{pred}}$, which is the weighted average of the models of the neighbors. At the end, it returns the prediction of $\boldy_{\text{pred}}$, which is the output of the model $\boldW_{\text{pred}}$ on the input data $\boldx$. 
We obtained the prediction by applying it to the input data $\boldx$.

\begin{algorithm}[t]
  \caption{Prediction with a new data point.\label{alg:pred}}
  \begin{algorithmic}
    \STATE {\bfseries Input:} Data $\boldx$
    \STATE $\boldx_{h_1}, .., \boldx_{h_K} \gets$ The $K$ nearest neighbors of $\boldx$ in the training set
    \STATE $\boldW_{\text{pred}} \gets 0$
    \STATE $S \gets 0$
    \FOR{$l \gets 1,\, n$}
    \STATE $c_l \gets 1/\norm{\boldy - \boldx}_2$
    \STATE $S \gets S + c_l$
    \STATE $\boldW_{\text{pred}} \gets \boldW_{\text{pred}} + c_l \boldW_{h_l}$
    \ENDFOR
    \STATE $\boldW_{\text{pred}} \gets \boldW_{\text{pred}}/S$
    \STATE $\boldy_{\text{pred}} \gets \boldW_{\text{pred}}^\top \boldx $
    \STATE {\bfseries Output:} $\boldy_{\text{pred}}$
  \end{algorithmic}
\end{algorithm}

Our method can be considered similar to the KNN approach because it is also an instance-based learning approach. However, instead of interpolating the values here in the $\mathbbR^m$ space, we interpolate the regularized models in the $\mathbbR^{d \times m}$ space. At this step, our model uses more input data than the KNN because $\boldx$ is again used here to compute the value of $\boldy$. Thus, if a new data point is significantly far from the input data, instead of predicting a value close to an already seen value, our model will attempt to linearly extend the prediction. 

The FALL method can be easily extended to new samples. For a new training sample $i_{\text{new}}$, only $\boldW_{i_{\text{new}}}$ needs to be computed. Thus, the training can be evolutive and there would be no need to compute any model again (in comparison to network Lasso for example).

\noindent Considering the final prediction for a new sample, we can also decompose the model predicted by $\boldW_{\text{pred}}$ using
\begin{align*}
    \boldW_{\text{pred}} &= \sum_{j=1}^{K_p} \alpha_{h_j} \left(\boldC_{h_j} + \boldA_{l_{h_j}} \right) \\
    &= \underbrace{\left[ \sum_{j=1}^{K_p} \alpha_{h_j} \boldC_{h_j} \right]}_{\boldC_{\text{pred}}} + \underbrace{\left[\sum_{j=1}^{K_p} \alpha_{h_j} \boldA_{l_{h_j}} \right]}_{\boldR_{\text{pred}}},
\end{align*}
where $\boldR_{\text{pred}}$ is the weighted mixture of the best anchor models of the neighborhood of $\boldx$ and $\boldC_{\text{pred}}$ is the weighted average of the correcting models. This correction term attempts to correct the fully regularized model by using the labels $\boldy_i$ to obtain a solution with a smaller training error. This term vanish when $\lambda \rightarrow \infty$.

\begin{algorithm}[t]
  \caption{Finding the anchor models.\label{alg:anc}}
  \begin{algorithmic}
    \STATE {\bfseries Input:} Data $(\boldx_1, \ldots,
    \boldx_n)$, ground truth $(\boldy_1, \ldots, \boldy_n)$
    \STATE $\boldx_{i_1}, \ldots, \boldx_{i_k} \gets$ Find anchor points (randomly or by K-means for example)
    \FOR{$l \gets 1,\, k$}
    \STATE $S_l \gets$ The neighbors set of $\boldx_{i_l}$ in $(\boldx_1, \ldots, \boldx_n)$
    \STATE $\boldA_l \gets$ Model computed with $\{ \boldx_{j} \}_{j \in S_l}$ and $\{ \boldy_{j} \}_{j \in S_l}$
    \ENDFOR
    \STATE {\bfseries Output:} $(\boldA_1, \ldots,
    \boldA_k)$
  \end{algorithmic}
\end{algorithm}

\subsection{Anchor model}
\label{sec:anchors}
The selection of anchors can be the subject of further research; however, we provide two simple methods for selecting them here.

First, by randomly taking a set of indexes $I = \{ i_1, \ldots, i_k\}$, we get our set of anchor points $\boldx_{i_1}, \ldots, \boldx_{i_k}$. Second, we can use a clustering algorithm like K-means to get $\boldx_{i_1}, \ldots, \boldx_{i_k}$. These vectors can be outside the set of input vectors.

These anchor points work as reference models in the training data. For each point, we select a set of neighbors (with $K$ nearest or some graph information) and compute a linear model (with Lasso \cite{JRSSB:Tibshirani:1996} or Ridge regression for example) (see Algorithm \ref{alg:anc}). Note that the for loop in Algorithm \ref{alg:anc} can be parallelized to speed up the process in Algorithm \ref{alg:anc}. The anchor points can be initialized using any method. We propose here a randomized method or a method with initialization using K-means clustering (and kmeans++).





\begin{figure}[t]
    \centering
    \includegraphics[scale=0.5]{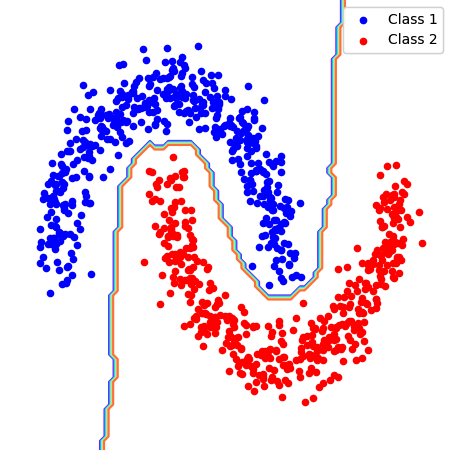}
    \caption{Two moons dataset. The boundary is established when the model predicts the same probability for the two classes. It can be observed that this boundary is linear by part and effectively separates the samples.}\label{fig:moons}
\end{figure}

\section{Experiments}
In this section, we first illustrate the proposed algorithm using synthetic datasets and then evaluate it using real-world datasets. 
\subsection{Illustrative experiments}
\begin{figure*}[t]
    \centering
    \subfigure[Synthetic dataset]{\label{fig:simple_a}\includegraphics[scale=0.35]{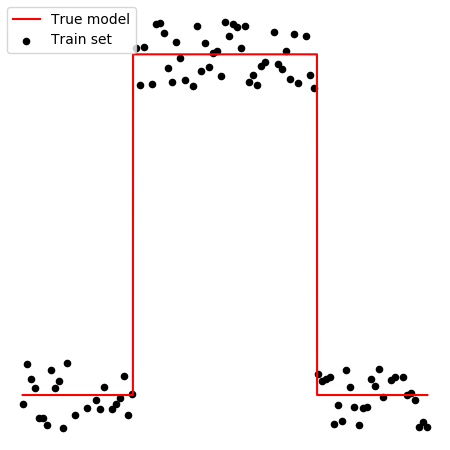}}
    \subfigure[Anchors models]{\label{fig:simple_b}\includegraphics[scale=0.35]{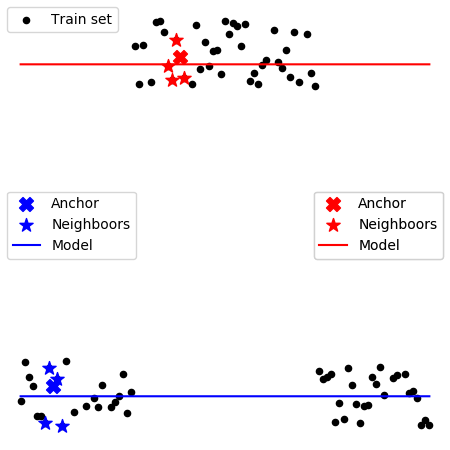}}
    \subfigure[Clustering]{\label{fig:simple_c}\includegraphics[scale=0.35]{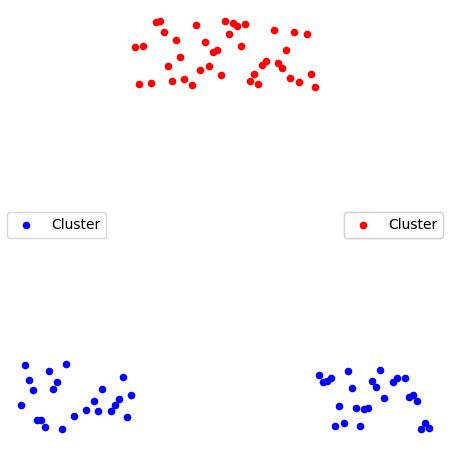}}
    \subfigure[Prediction of our model with various $\lambda$]{\label{fig:simple_d}\includegraphics[scale=0.35]{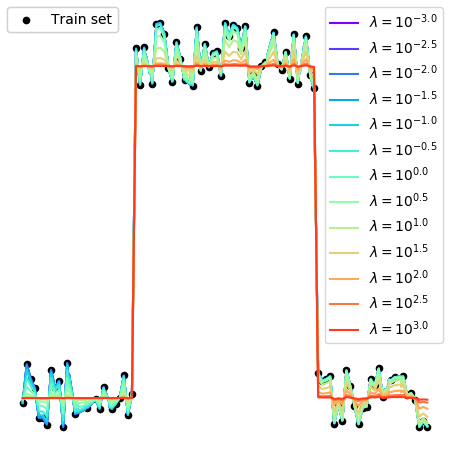}}
    \caption{Generation of synthetic dataset in Figure \ref{fig:simple_a} (a step function), which is non linear. The red plot is the real function and the black dots are the training set, which is generated with some added noise.
    In Figure \ref{fig:simple_b}, we can see the anchor points with their relative neighborhood and computed anchor models. They are computed with the linear Ridge regression.
    In Figure \ref{fig:simple_c}, the data points are clustered into $k$ clusters by choosing the best anchor model to model the linear relationship between a sample $\boldx_i$ and its label $\boldy_i$ (the best $\boldA_l$ that minimizes $||\boldy_i - \boldA_l ^\top \boldx_i||$). The points on the right have the same behavior locally than those on the left.
    Finally, Figure \ref{fig:simple_d} shows our computed final model depending on the $\lambda$ regularization parameter. The continuous path between the "perfect" solution, where each model predicts the correct label, and the most regularized solution, where locally it is the same as the model of the best anchor at this position, can be observed.}\label{fig:simple}
    
\end{figure*}

\vspace{.05in}
\noindent {\bf Step function:}
To present our method and demonstrate its working, let us consider a simple non linear model, i.e., step function, which only takes two values. Let us sample some random points with some random noise (see in Figure \ref{fig:simple_a}).

Let us take $k=2$ anchors points (we could take more, this is for illustration), one inside and one outside. There are parts of the data where the linear models are not similar. Let us compute their nearest neighbors and anchor models for each neighborhood (as shown in Figure \ref{fig:simple_b}). The models are determined by applying ridge regression with bias, with a high regularization parameter $\alpha = 1000$.

For each data point $\boldx_i$, we search for the anchor model $\boldA_{l_i}$ which has the minimum loss $\norm{\boldy_i - \boldA_{l_i} ^\top \boldx_i}$. This is the best linear anchor model that explains the relation between $\boldx_i$ and $\boldy_i$ (and corresponds to selecting the optimal $\boldp_i = \bolddelta_{l_i}$). This provides a natural method to cluster our data into $k$ clusters (see Figure \ref{fig:simple_c}).

Note that the proposed FALL method can cluster data points with respect to anchor models (clustering by regression), while it is not possible to directly cluster points using KNN and other linear models. Thus, the proposed FALL model has a similar paradigm as network Lasso.

\noindent For each $\boldx_i$, after the clustering step (finding the best anchor model $\boldA_l$), we compute the final model $\widehat{\boldW}_i$ by 
\begin{displaymath}
\widehat{\boldW}_i = \frac{1}{\lambda + \boldx_i ^\top \boldx_i} \boldx_i \left[\boldy_i  - \boldA_l ^\top \boldx_i \right] ^\top + \boldA_l.
\end{displaymath}
As evident in Figure \ref{fig:simple_d}, larger values of $\lambda$ imply that the final model is farther from the training data but closer to the anchor models.

\vspace{.05in}
\noindent {\bf Two moons dataset:}
The proposed method can also be applied to the classification datasets. If the model is trained with one-hot encoding vectors $\boldy$ (i.e. when $\boldy = \delta_j$), our prediction can be interpreted as probability vectors. To illustrate this, we classify the two moons dataset by using Eq.~\eqref{eq:eq3}.


In Figure \ref{fig:moons}, boundaries can also be observed (the contour when the probability to be in each class is equal). The proposed method can separate the two moons very well in this example.
                     
\begin{table*}[t]
\small
\setlength\tabcolsep{0pt}
\caption{Averaged mean squared error on the real-world datasets over 50 runs. For each method, we select the best parameters by using 3-fold cross-validation. N/A means that a memory problem occurred when computing the model on the dataset. The asterisk * means that the results were computed without cross-validation but with fixed parameters (because of a too long runtime).}
\begin{tabular*}{\textwidth}{@{\extracolsep{\fill}} >{\bfseries}l *{7}{c}}
\toprule
{} &                     FALL  &                 Network Lasso  &                KNN &                 KRR &               Ridge &               Lasso \\
\midrule
Concrete data             &  42.16 ($\pm$ 19.38) &  49.85 ($\pm$ 22.12)  &  58.14 ($\pm$ 26.85) &  110.80 ($\pm$ 33.57) &  111.05 ($\pm$ 33.92) &  111.10 ($\pm$ 33.88) \\
Superconduct data            & 129.66 ($\pm$ 28.33)   & N/A    &  117.48 ($\pm$ 18.77) &    312.11 ($\pm$ 25.82) &  308.52 ($\pm$ 22.56) & 345.25 ($\pm$ 99.80) \\
Aquatic toxicity data   & 1.46 ($\pm$ 0.74) &  1.72 ($\pm$ 1.01) &  1.81 ($\pm$ 0.74) &  1.92 ($\pm$ 0.75) & 1.58 ($\pm$ 0.72) &  1.58 ($\pm$ 0.71) \\
Fish toxicity data &  0.92 ($\pm$ 0.48) &   0.98 ($\pm$ 0.53) &  0.91 ($\pm$ 0.48) &  1.14 ($\pm$ 0.57) &  0.99 ($\pm$ 0.44) &  0.99 ($\pm$ 0.44) \\
P.T. motor UPDRS data  &  20.76 ($\pm$ 4.51) & 24.63 ($\pm$ 6.89)* & 20.88 ($\pm$ 4.29) &  58.30 ($\pm$ 5.50) &  56.69 ($\pm$ 5.26) &  56.81 ($\pm$ 5.20) \\
\bottomrule
\end{tabular*}
\vspace{-.1in}
\end{table*}


\begin{table*}
\small
\setlength\tabcolsep{0pt}
\caption{Training time and testing time (in seconds) on real-world datasets over 50 runs. For FALL, the training time is the sum of the creation of the anchor models and the final model. N/A means that a memory problem occurred when computing the model on the dataset. The asterisk * means that the training time and testing time were averaged with few samples (because of a too long runtime).}
\begin{tabular*}{\textwidth}{@{\extracolsep{\fill}} >{\bfseries}l *{8}{c}}
\toprule
{} &  &                                      FALL  &                 Network Lasso  &                KNN &                 KRR &               Ridge &               Lasso \\
\midrule
\multirow{2}{*}{Concrete data}  & Train      &  0.06 ($\pm$ 0.05)  &    16.58 ($\pm$ 3.41)  &    0.00 ($\pm$ 0.00) &     0.09 ($\pm$ 0.08) &     0.00 ($\pm$ 0.00) &     0.01 ($\pm$ 0.00) \\
& Test      &   0.01 ($\pm$ 0.00)  &    0.19 ($\pm$ 0.03)  &    0.00 ($\pm$ 0.00) &     0.01 ($\pm$ 0.02) &     0.00 ($\pm$ 0.00) &     0.00 ($\pm$ 0.00) \\
\midrule
\multirow{2}{*}{Superconduct data}  &  Train  & 1.24 ($\pm$ 1.33)  & N/A  &     0.11 ($\pm$ 0.02) &     124.61 ($\pm$ 109.20) &     0.02 ($\pm$ 0.01) &   121.04 ($\pm$ 125.75) \\
& Test & 5.90 ($\pm$ 0.71) & N/A  &     0.22 ($\pm$ 0.08) &      0.23 ($\pm$ 0.13) &   0.00 ($\pm$ 0.01) &       0.01 ($\pm$ 0.01) \\
\midrule
\multirow{2}{*}{Aquatic toxicity data} & Train & 0.02 ($\pm$ 0.05) &  5.31 ($\pm$ 1.00) &  0.00 ($\pm$ 0.00) &  0.01 ($\pm$ 0.01) & 0.00 ($\pm$ 0.00) &  0.00 ($\pm$ 0.00) \\
& Test & 0.00 ($\pm$ 0.00) &  0.07 ($\pm$ 0.01) &  0.00 ($\pm$ 0.00) &  0.00 ($\pm$ 0.01) & 0.00 ($\pm$ 0.00) &  0.00 ($\pm$ 0.00) \\
\midrule
\multirow{2}{*}{Fish toxicity data} & Train & 0.05 ($\pm$ 0.05) &  13.23 ($\pm$ 1.78) &  0.00 ($\pm$ 0.00) &  0.03 ($\pm$ 0.04) &  0.00 ($\pm$ 0.00) &  0.00 ($\pm$ 0.00) \\
& Test &  0.01 ($\pm$ 0.00) &   0.14 ($\pm$ 0.02) &  0.00 ($\pm$ 0.00) &  0.00 ($\pm$ 0.01) &  0.00 ($\pm$ 0.00) &  0.00 ($\pm$ 0.00) \\
\midrule
\multirow{2}{*}{P.T. motor UPDRS data} & Train &  0.25 ($\pm$ 0.99) & $\sim$ 30 min* &  0.00 ($\pm$ 0.00) &   1.39 ($\pm$ 0.51) &   0.00 ($\pm$ 0.00) &   0.01 ($\pm$ 0.01) \\
 & Test &  0.27 ($\pm$ 0.02) & $\sim$ 4 seconds* &  0.01 ($\pm$ 0.00) &   0.02 ($\pm$ 0.01) &   0.00 ($\pm$ 0.01) &   0.01 ($\pm$ 0.01) \\
\bottomrule
\end{tabular*}
\vspace{-.1in}
\end{table*}

\subsection{Real-world dataset}
\noindent {\bf Setup:} We compared our method on several datasets with other methods. We computed the average mean squared error with 50 runs and the training and test time. 
We compared the proposed FALL model with KNN, kernel ridge regression (KRR), ridge and Lasso \cite{tibshirani2005sparsity}, and network Lasso \cite{hallac2015network}. It should be noted that network Lasso is a state-of-the-art locally linear model. To compare the methods fairly, we selected the best hyperparameters by cross-validation with optuna \cite{optuna_2019} with a timeout of $60$ s. 

\noindent For each dataset, the following procedure was performed.
\begin{enumerate}
    \item Randomly split the dataset into training and testing data with a proportion of $0.9/0.1$.
    \item Tune the hyperparameters using the 3-cross-validation score on the training data ($60$ s maximum for tunning).
    \item Use these parameters to fit the entire training set and time it.
    \item Test the model on the test set and time it.
    \item Repeat (1) 50 times and average the results.
\end{enumerate}

\noindent All splits are the same for each method.

\vspace{.05in}
\noindent {\bf Concrete Compressive Strength Dataset:} In civil engineering, concrete is a widely used material. For high-performance concrete, the compressive strength is a highly nonlinear function of age and ingredients. This information is useful to understand the importance of each ingredient to create more resistant concrete. In this dataset \cite{concrete}, the number of attributes (dimension) is $d=9$ and the number of samples is $n=1030$.


\vspace{.05in}
\noindent {\bf Superconductivty Dataset:} Health care professionals or researchers in the Hadron Collider at CERN are interested in superconducting materials. These materials are interesting because they are conductive without any resistance. To use this property, superconducting materials must be used in their critical temperature, which can be difficult to find. 
This dataset \cite{HAMIDIEH2018346} contains $d=81$ interesting features (which are related to the chemical formula) and the goal is to predict the critical temperature. There are $n=21263$ samples in this dataset. We could not use network Lasso due to memory error (too large matrix to perform factorization).
      

\vspace{.05in}
\noindent {\bf QSAR Toxicity Datasets:} The water ecosystem can be damaged by some chemicals. These two datasets QSAR aquatic toxicity \cite{aquatictoxicitydataset} and fish toxicity \cite{fishtoxicitydataset} were created to understand what combinations of chemicals that can be highly toxic for certain organisms. The value of toxicity to be predicted is the concentration of LC50 in Daphnia magna (a small planktonic crustacean). This is a good indicator of toxicity because it kills this creature. The difference between the two datasets is that the aquatic toxicity dataset contains $d=8$ features (which corresponds to eight molecular descriptors), $n=546$ instances and the output is the concentration of LC50 after 48h; the fish toxicity dataset contains $d=6$ features, $n=908$ instances and the output is the concentration of LC50 after 96h.


\vspace{.05in}
\noindent {\bf Parkinsons Telemonitoring Data Set:}  The symptoms of Parkinson's disease need to be monitored by a doctor in a clinic, which is costly and time-consuming. They can use the unified Parkinson's disease rating scale (UPDRS) to track the progression of the symptoms. This dataset \cite{parkinsontele} contains information about the patients (as age or gender) and $16$ biomedical voice measures, which are taken in the patient home with a telemonitoring device, to form $d=19$ features. There are $n=5875$ instances and one of the interesting output is the motor UPDRS. This target value can help people with early-stage Parkinson's disease to follow up more conveniently and less expensively. 

\vspace{.05in}
\noindent {\bf Result:} Through experiments, the proposed FALL method was found to compare favorably with network Lasso and other existing baselines. In particular, the concrete data, the FALL method significantly outperforms baseline methods. Overall, because the FALL method is a locally linear model, the performance of the FALL method is similar to the one with KNN. The important point here is that the computation time of the FALL method is two orders of magnitude smaller than that of network Lasso.

\section{Conclusion}

Local linear methods are nonlinear methods that can yield good results on very complex data. In this study, we propose a new locally linear model called the fast anchor regularized local linear method (FALL), which produces a formula that can be used to efficiently compute a regularized locally linear model. The FALL method also enables us to understand the mechanism of regularization and interpret it through its inner clustering. Through experiments, we demonstrated that the proposed method compares favorably with network Lasso, which is a state-of-the-art locally linear model. Moreover, the computation time of FALL model is two orders of magnitude smaller than that of network Lasso; additionally, this model can be extended with new training data with a small computational cost.

\bibliography{main_arxiv}

\begin{thebibliography}{10}

\bibitem{optuna_2019}
Takuya Akiba, Shotaro Sano, Toshihiko Yanase, Takeru Ohta, and Masanori Koyama.
\newblock Optuna: A next-generation hyperparameter optimization framework.
\newblock In {\em KDD}, 2019.

\bibitem{bartlett1951}
M.~S. Bartlett.
\newblock An inverse matrix adjustment arising in discriminant analysis.
\newblock {\em Ann. Math. Statist.}, 22(1):107--111, 03 1951.

\bibitem{fishtoxicitydataset}
M.~Cassotti, D.~Ballabio, R.~Todeschini, and V.~Consonni.
\newblock A similarity-based qsar model for predicting acute toxicity towards
  the fathead minnow (pimephales promelas).
\newblock {\em SAR and QSAR in Environmental Research}, 26(3):217--243, 2015.

\bibitem{aquatictoxicitydataset}
Matteo Cassotti, Davide Ballabio, Viviana Consonni, Andrea Mauri, Igor Tetko,
  and Roberto Todeschini.
\newblock Prediction of acute aquatic toxicity toward daphnia magna by using
  the ga-knn method.
\newblock {\em Alternatives to laboratory animals : ATLA}, 42:31--41, 03 2014.

\bibitem{knn}
T.~{Cover} and P.~{Hart}.
\newblock Nearest neighbor pattern classification.
\newblock {\em IEEE Transactions on Information Theory}, 13(1):21--27, 1 1967.

\bibitem{agepred}
Jason~G. Fleischer, Roberta Schulte, Hsiao~H. Tsai, Swati Tyagi, Arkaitz
  Ibarra, Maxim~N. Shokhirev, Ling Huang, Martin~W. Hetzer, and Saket Navlakha.
\newblock Predicting age from the transcriptome of human dermal fibroblasts.
\newblock {\em Genome Biology}, 19(1):221, 2018.

\bibitem{housepred}
Guangliang Gao, Zhifeng Bao, Jie Cao, A.~Kai Qin, Timos Sellis, and Zhiang Wu.
\newblock Location-centered house price prediction: {A} multi-task learning
  approach.
\newblock {\em CoRR}, abs/1901.01774, 2019.

\bibitem{hallac2015network}
David Hallac, Jure Leskovec, and Stephen Boyd.
\newblock Network lasso: Clustering and optimization in large graphs.
\newblock In {\em KDD}, 2015.

\bibitem{hallac2015snapvx}
David Hallac, Christopher Wong, Steven Diamond, Rok Sosic, Stephen Boyd, and
  Jure Leskovec.
\newblock Snapvx: A network-based convex optimization solver.
\newblock {\em arXiv preprint arXiv:1509.06397}, 2015.

\bibitem{HAMIDIEH2018346}
Kam Hamidieh.
\newblock A data-driven statistical model for predicting the critical
  temperature of a superconductor.
\newblock {\em Computational Materials Science}, 154:346 -- 354, 2018.

\bibitem{agepredimg}
Hu~Han, Charles Otto, and Anil Jain.
\newblock Age estimation from face images: Human vs. machine performance.
\newblock 06 2013.

\bibitem{finance2}
Lucas~Cassiel Jacaruso.
\newblock A method of trend forecasting for financial and geopolitical data:
  inferring the effects of unknown exogenous variables.
\newblock {\em Journal of Big Data}, 5(1):47, 12 2018.

\bibitem{scikit-learn}
F.~Pedregosa, G.~Varoquaux, A.~Gramfort, V.~Michel, B.~Thirion, O.~Grisel,
  M.~Blondel, P.~Prettenhofer, R.~Weiss, V.~Dubourg, J.~Vanderplas, A.~Passos,
  D.~Cournapeau, M.~Brucher, M.~Perrot, and E.~Duchesnay.
\newblock Scikit-learn: Machine learning in {P}ython.
\newblock {\em Journal of Machine Learning Research}, 12:2825--2830, 2011.

\bibitem{peyre2019computational}
Gabriel Peyr{\'e}, Marco Cuturi, et~al.
\newblock Computational optimal transport.
\newblock {\em Foundations and Trends{\textregistered} in Machine Learning},
  11(5-6):355--607, 2019.

\bibitem{sentiment2}
Colin Raffel, Noam Shazeer, Adam Roberts, Katherine Lee, Sharan Narang, Michael
  Matena, Yanqi Zhou, Wei Li, and Peter~J. Liu.
\newblock Exploring the limits of transfer learning with a unified text-to-text
  transformer, 2019.

\bibitem{book:Schoelkopf+Smola:2002}
B.~Sch\"olkopf and A.~J. Smola.
\newblock {\em Learning with Kernels}.
\newblock MIT Press, Cambridge, MA, 2002.

\bibitem{sherman1950}
Jack Sherman and Winifred~J. Morrison.
\newblock Adjustment of an inverse matrix corresponding to a change in one
  element of a given matrix.
\newblock {\em Ann. Math. Statist.}, 21(1):124--127, 03 1950.

\bibitem{sentiment1}
Tan Thongtan and Tanasanee Phienthrakul.
\newblock Sentiment classification using document embeddings trained with
  cosine similarity.
\newblock In {\em Proceedings of the 57th Annual Meeting of the Association for
  Computational Linguistics: Student Research Workshop}, pages 407--414,
  Florence, Italy, July 2019. Association for Computational Linguistics.

\bibitem{lasso}
R.~Tibshirani.
\newblock Regression shrinkage and selection via the lasso.
\newblock {\em Journal of the Royal Statistical Society (Series B)},
  58:267--288, 1996.

\bibitem{JRSSB:Tibshirani:1996}
R.~Tibshirani.
\newblock Regression shrinkage and selection via the {L}asso.
\newblock {\em Journal of the Royal Statistical Society, Series B},
  58(1):267--288, 1996.

\bibitem{tibshirani2005sparsity}
Robert Tibshirani, Michael Saunders, Saharon Rosset, Ji~Zhu, and Keith Knight.
\newblock Sparsity and smoothness via the fused lasso.
\newblock {\em Journal of the Royal Statistical Society: Series B (Statistical
  Methodology)}, 67(1):91--108, 2005.

\bibitem{parkinsontele}
Athanasios Tsanas, Max Little, Patrick Mcsharry, and Lorraine Ramig.
\newblock Accurate telemonitoring of parkinson's disease progression by
  noninvasive speech tests.
\newblock {\em IEEE transactions on bio-medical engineering}, 57:884--93, 11
  2009.

\bibitem{yamada2016localized}
Makoto Yamada, Koh Takeuchi, Tomoharu Iwata, John Shawe-Taylor, and Samuel
  Kaski.
\newblock Localized lasso for high-dimensional regression.
\newblock In {\em AISTATS}, 2017.

\bibitem{concrete}
I-Cheng Yeh.
\newblock Modeling of strength of high-performance concrete using artificial
  neural networks.” cement and concrete research, 28(12), 1797-1808.
\newblock {\em Cement and Concrete Research}, 28:1797--1808, 12 1998.

\bibitem{finance1}
Yu-Dong Zhang and Lenan Wu.
\newblock Stock market prediction of sp 500 via combination of improved bco
  approach and bp neural network.
\newblock {\em Expert Systems with Applications}, 36:8849--8854, 07 2009.

\bibitem{zou2005regularization}
H.~Zou and T.~Hastie.
\newblock Regularization and variable selection via the elastic net.
\newblock {\em Journal of the Royal Statistical Society: Series B (Statistical
  Methodology)}, 67(2):301--320, 2005.

\end{thebibliography}
\bibliographystyle{plain}



\clearpage
\appendix
\section*{Supplementary material}
\subsection*{Implementation and searching the hyper-parameters}

For each method, the implementation used for the experiments and the hyper-parameter space are explained for the optimization using optuna \cite{optuna_2019}.

\subsubsection*{Proposed method}
Our implementation will be available online soon. It was implemented in Python 3 with numpy and to yield a "sklearn" like model.
\begin{itemize}
\item The number of anchor points \\ $k \in \{ 20, 40, 60, 80, 100, 120, 140, 160, 180, 200\}$
\item The number of neighbors to create the anchor model \\ $K_{\text{anchors}} \in \{ 5, 10, 15, 20, 25, 30, 35, 40, 45, 50\}$
\item Use bias inside the models: True or False
\item Regularization for the anchor models \\ $\alpha \in \{ 10^{-3}, 10^{-2}, 10^{-1}, 10^{0}, 10^{1}, 10^{2}, 10^{3}\}$
\item The l1-ratio in Elastic net for the anchor models \\ $R \in \{ 0.0, 0.1, 0.2, 0.3, 0.4, 0.5, 0.6, 0.7, 0.8, 0.9, 1.0\}$
\item The regularization parameter \\ $\lambda \in \{ 10^{-3}, 10^{-2}, 10^{-1}, 10^{0}, 10^{1}, 10^{2}, 10^{3}\}$
\item The number of neighbors for the prediction \\ $K_{\text{pred}} \in \{ 5, 10, 15, 20, 25, 30, 35, 40, 45, 50\}$
\end{itemize}

\subsubsection*{Network Lasso}
We used the Python implementation of network Lasso following the work by the authors of Localized lasso \cite{yamada2016localized}. It solves the same problem when we set $\lambda_{\text{exc}} = 0$. We tested it with $K=5$ neighbors to construct the graph. 

\begin{itemize}
\item The regularization parameter \\ $\lambda_{\text{net}} \in \{ 10^{-3}, 10^{-2}, 10^{-1}, 10^{0}, 10^{1}, 10^{2}, 10^{3}\}$
\item Use bias inside the models: True or False
\end{itemize}

There are fewer parameters here, but due to the restriction in training time, we cannot add more parameters. 

\subsubsection*{Kernel Ridge Regression}
We used the scikit-learn \cite{scikit-learn} implementation of KRR.

\begin{itemize}
\item The regularization parameter \\ $\alpha \in \{ 10^{-3}, 10^{-2}, 10^{-1}, 10^{0}, 10^{1}, 10^{2}, 10^{3}\}$
\item The kernel used: RBF or poly
\item Parameters for RBF kernel \begin{itemize}
    \item The inverse of the standard deviation of the RBF kernel \\ $\gamma \in \{ 10^{-3}, 10^{-2}, 10^{-1}, 10^{0}, 10^{1}, 10^{2}, 10^{3}\}$
\end{itemize}
\item Parameters for poly kernel \begin{itemize}
    \item The degree of the polynomial \\ $d \in \{ 1, 2, 3, 4, 5, 6, 7, 8, 9, 10 \}$
    \item Zero coefficient \\ $c \in \{ 10^{-3}, 10^{-2}, 10^{-1}, 10^{0}, 10^{1}, 10^{2}, 10^{3}\}$
\end{itemize}
\end{itemize}

\subsubsection*{K-Nearest Neighbors (KNN)}
We used the scikit-learn \cite{scikit-learn} implementation of KNN.
\begin{itemize}
\item The number of neighbors \\ $K \in \{ 5, 10, 15, 20, 25, 30, 35, 40, 45, 50 \}$
\item Weights: Uniform or Weights (use inverse distance like our approach)
\end{itemize}

\subsubsection*{Ridge or Lasso}
We used the scikit-learn \cite{scikit-learn} implementation of Ridge and Lasso.
                
\begin{itemize}
\item The regularization parameter \\ $\alpha \in \{ 10^{-3}, 10^{-2}, 10^{-1}, 10^{0}, 10^{1}, 10^{2}, 10^{3}\}$
\item The number of neighbors \\ $K \in \{ 5, 10, 15, 20, 25, 30, 35, 40, 45, 50 \}$
\item Use bias inside the models: True or False
\end{itemize}

\newpage
\subsection*{Proofs of lemma}
\theoproof{\ref{lemm:quad}}
Let us rewrite the problem to show that it is a quadratic problem.

\begin{displaymath}
J(\boldP_i, \widehat{\boldW}_i) = \loss(\widehat{\boldW}_i) + \lambda \optloss(\boldP_i, \widehat{\boldW}_i)
\end{displaymath}

Let us focus on each term:

\vspace{.1in}
\noindent {\bf Mean square error term} 
\begin{align*}
  \loss(\hat{\boldW}_i) &= \norm{\boldy_i - \hat{\boldW}_i ^\top \boldx_i}^2 \\
  &= \norm{\boldy_i - \frac{\boldx_i^\top \boldx_i}{\lambda + \boldx_i ^\top \boldx_i} \boldy_i + \frac{\lambda}{\lambda + \boldx_i ^\top \boldx_i} \boldG_i^\top \boldx_i}^2 \\
    & = \frac{\lambda ^2}{(\lambda + \boldx_i ^\top \boldx_i)^2} \norm{\boldy_i - \boldG_i ^\top \boldx_i}^2
\end{align*}

\vspace{.1in}
\noindent {\bf Regularization term}
First, we will decompose the square norm of difference as a sum of norms and scalar product. 
\begin{align*}
  \optloss(\boldP_i, \hat{\boldW}_i) &= \sum_{l=1}^{k} p_{il} \norm{\widehat{\boldW}_i - \boldA_l}^2_F \\ 
  &= \sum_{l=1}^k p_{il} \left[ \norm{\widehat{\boldW}_i}^2_F - 2 \inner{\widehat{\boldW}_i}{\boldA_l}_F + \norm{\boldA_l}_F^2 \right]\\
  &= \norm{\widehat{\boldW}_i}_F^2 - 2 \inner{\widehat{\boldW}_i}{\boldG_i}_F + \sum_{l=1}^{k} p_{il} \norm{\boldA_l}_F^2
\end{align*}

\noindent Let us focus on the first term of $\optloss$ by injecting $\widehat{\boldW}_i$.
\begin{align*}
  \norm{\widehat{\boldW}_i}_F^2 &= \norm{\frac{1}{\lambda + \boldx_i ^\top \boldx_i} \boldx_i \left[\boldy_i  - \boldG_i ^\top \boldx_i \right]^\top + \boldG_i}_F^2 \\
                                  &= \frac{1}{(\lambda + \boldx_i ^\top \boldx_i)^2} \norm{\boldx_i \left[\boldy_i  - \boldG_i ^\top \boldx_i \right]^\top}_F^2 + 2 \inner{\frac{1}{\lambda + \boldx_i ^\top \boldx_i} \boldx_i \left[\boldy_i  - \boldG_i ^\top \boldx_i \right]}{\boldG_i}_F+  \norm{\boldG_i}_F^2
\end{align*}

\noindent The first norm of this expression can be rewritten as
\begin{align*}
  \norm{\boldx_i \left[\boldy_i  - \boldG_i ^\top \boldx_i \right]^\top}_F^2 &= \tra(\left[\boldy_i  - \boldG_i ^\top \boldx_i \right]\boldx_i^\top \boldx_i \left[\boldy_i  - \boldG_i ^\top \boldx_i \right]^\top)  \\
  &= \boldx_i ^\top \boldx_i \tra(\left[\boldy_i  - \boldG_i ^\top \boldx_i \right]\left[\boldy_i  - \boldG_i ^\top \boldx_i \right]^\top) \\
  &= \boldx_i ^\top \boldx_i \norm{\boldy_i  - \boldG_i ^\top \boldx_i}_2^2
\end{align*}

\noindent Let us focus on the second term of $\optloss$ by injecting $\widehat{\boldW}_i$.
\begin{align*}
  \inner{\widehat{\boldW}_i}{\boldG_i}_F &= \inner{\frac{1}{\lambda + \boldx_i ^\top \boldx_i} \boldx_i \left[\boldy_i  - \boldG_i ^\top \boldx_i \right]^\top + \boldG_i}{\boldG_i}_F \\
                                         &= \inner{\frac{1}{\lambda + \boldx_i ^\top \boldx_i} \boldx_i \left[\boldy_i  - \boldG_i ^\top \boldx_i \right]^\top}{\boldG_i}_F + \norm{\boldG_i}^2_F
\end{align*}

\noindent The inner product term gets cancelled. Thus, we obtain the following formula.
\begin{align*}
  \optloss(\boldP_i, \widehat{\boldW}_i) &= \frac{ \boldx_i ^\top \boldx_i}{(\lambda + \boldx_i ^\top \boldx_i)^2} \norm{\boldy_i - \boldG_i^\top \boldx_i}_2^2 + \sum_{l=1}^{k} p_{il} \norm{\boldA_l}_F^2 - \norm{\boldG_i}_F^2
\end{align*}

\noindent {\bf Combining the terms and factorization}

\noindent Let us inject these intermediate derivations into our problem.
\begin{align*}
  J(\boldP_i, \widehat{\boldW}_i) &= \loss(\widehat{\boldW}_i) + \lambda \optloss(\boldP_i, \widehat{\boldW}_i) \\
                                  &= \frac{1}{(\lambda + \boldx_i ^\top \boldx_i)^2} \norm{\boldy_i - \boldG_i^\top \boldx_i}_2^2 (\lambda^2 + \lambda \boldx_i ^\top \boldx_i) \lambda \left( \sum_{l=1}^{k} p_{il} \norm{\boldA_l}_F^2 - \norm{\boldG_i}^2_F \right)
\end{align*}

\noindent The first term $T_1$ can be rewritten as
\begin{align*}
T_1 &= \beta_i \norm{\boldy_i - \boldG_i^\top \boldx_i}_2^2 \\
    &= \beta_i \left( \norm{\boldy_i}_2^2 - 2 \boldy_i ^\top \boldG_i^\top \boldx_i + \norm{\boldG_i^\top \boldx_i}_2^2 \right) \\
    &= \beta_i \left( \norm{\boldy_i}_2^2 - 2 \sum_{l=1}^{k} p_{il} \boldy_i^\top \boldA_l^\top \boldx_i + \sum_{l=1}^{k} \sum_{l^{\prime}=1}^{k} p_{il} p_{il^{\prime}} \boldx_i ^\top \boldA_l \boldA_{l^{\prime}} ^\top \boldx_i \right)
\end{align*}

\noindent The second term $T_2$ can be rewritten as
\begin{align*}
T_2 &= \lambda \left( \sum_{l=1}^{k} p_{il} \norm{\boldA_l}_F^2 - \norm{\boldG_i}^2_F \right) \\
    &= \lambda \left( \sum_{l=1}^{k} p_{il} \norm{\boldA_l}_F^2 - \sum_{l=1}^{k} \sum_{l^{\prime}=1}^{k} p_{il} p_{il^{\prime}} \tra(\boldA_l^\top \boldA_{l^{\prime}}) \right)
\end{align*}

\noindent Finally, when we combine the sum in $T_1$ and $T_2$, we get
\begin{align*}
J(\boldP_i, \widehat{\boldW}_i) &= \sum_{l=1}^{k} \sum_{l^{\prime}=1}^{k} p_{il} p_{il^{\prime}} \underbrace{(\beta_i \boldx_i ^\top \boldA_l \boldA_{l^{\prime}} ^\top \boldx_i - \lambda \tra(\boldA_l^\top \boldA_{l^{\prime}}))}_{(\boldH_i)_{ll^{\prime}}} + \sum_{l=1}^{k} p_{il} \underbrace{(-2 \beta_i \boldy_i ^ \top \boldA_l ^\top \boldx_i + \lambda \norm{\boldA_l}_F^2)}_{(\boldb_i)_l} + \beta_i \norm{\boldy_i}_2^2 \\
  &= \boldp_i ^\top \boldH_i \boldp_i + \boldb_i ^\top \boldp_i + \beta_i \norm{\boldy_i}_2^2
\end{align*}
  
\noindent Thus, the final function to be minimized is quadratic.

\proofend

\end{document}